\documentclass[conference]{IEEEtran}
\IEEEoverridecommandlockouts

\usepackage{paper}
\usepackage[utf8]{inputenc}
\usepackage{newunicodechar}
\newunicodechar{ }{\,}
\def\BibTeX{{\rm B\kern-.05em{\sc i\kern-.025em b}\kern-.08em
    T\kern-.1667em\lower.7ex\hbox{E}\kern-.125emX}}
\begin{document}

% \title{STONK: Stock Optimization using News Knowledge\\
% }
\title{Towards Unified Multimodal Financial Forecasting: Integrating Sentiment Embeddings and Market Indicators via Cross-Modal Attention}
\author{
\IEEEauthorblockN{Sarthak Khanna\IEEEauthorrefmark{2}\IEEEauthorrefmark{1}, Armin Berger\IEEEauthorrefmark{9}\IEEEauthorrefmark{2}\IEEEauthorrefmark{3}\IEEEauthorrefmark{1}, 
David Berghaus\IEEEauthorrefmark{9}, \\ Tobias Deusser\IEEEauthorrefmark{9}\IEEEauthorrefmark{2}, Lorenz Sparrenberg\IEEEauthorrefmark{2}, Rafet Sifa\IEEEauthorrefmark{9}\IEEEauthorrefmark{2}}
\IEEEauthorblockA{\IEEEauthorrefmark{9}Fraunhofer IAIS - Department of Media Engineering, Germany}
\IEEEauthorblockA{\IEEEauthorrefmark{2}University of Bonn - Department of Computer Science, Germany}
\IEEEauthorblockA{\IEEEauthorrefmark{3}West-AI - Federal Ministry of Education and Research, Germany}

\thanks{* Both authors contributed equally to this research.}\\
\thanks{The project was funded by the Federal Ministry of Education and Research (BMBF) under grant no. 01IS22094A WEST-AI. This research has been partially funded by the Federal Ministry of Education and Research of Germany and the state of North-Rhine Westphalia as part of the Lamarr-Institute for Machine Learning and Artificial Intelligence.}
}

\maketitle
\begin{abstract}
We propose STONK (Stock Optimization using News Knowledge), a multimodal framework integrating numerical market indicators with sentiment-enriched news embeddings to improve daily stock-movement prediction. By combining numerical \& textual embeddings via feature concatenation and cross-modal attention, our unified pipeline addresses limitations of isolated analyses. Backtesting shows STONK outperforms numeric-only baselines. A comprehensive evaluation of fusion strategies and model configurations offers evidence-based guidance for scalable multimodal financial forecasting. Source code is available on GitHub\footnote{\url{https://github.com/sarthak-12/thesis-dsaa/}}.

\end{abstract}


\section{Introduction}

The intersection of financial markets and AI has expanded dramatically, driven by advances in modeling techniques and data availability. Traditional approaches relied on numerical indicators while neglecting contextual information in financial news. Markets are complex systems influenced by multiple factors, and previous approaches typically separated textual and numerical analysis or used simplistic sentiment scoring.

Lopez-Lira et al.~\cite{lopez-lira2025memorization} demonstrate that the apparent forecasting ability of state-of-the-art Large Language Models (LLMs) often reflects memorization rather than genuine prediction. To address these limitations, we introduce STONK (Stock Optimization using News Knowledge), a multimodal framework integrating news with market indicators through concatenation and cross-modal attention.

Our key contributions are: (1) multimodal fusion of numerical embeddings and sentiment-annotated textual features; (2) domain-adaptive sentiment scoring via RoBERTa for precise market signals; and (3) a robust backtesting framework with rolling time-series splits and financial metrics for model validation.

\begin{figure}[ht]
    \centering
    \includegraphics[width=5.5cm]{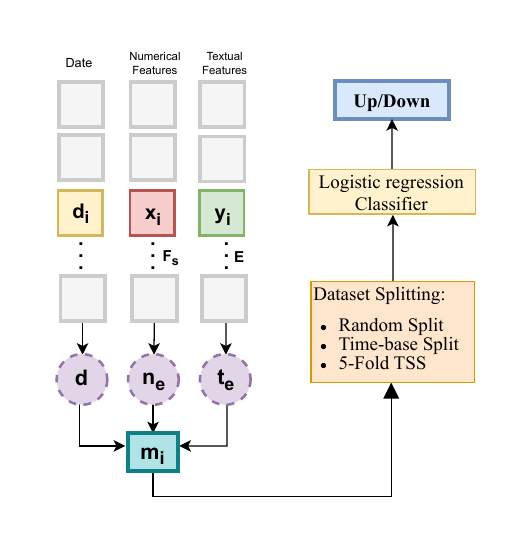}
        \caption{Multimodal pipeline for sample $i$. Numeric features $x_i$ and textual
  embedding $y_i$ are encoded and fused into the market vector $m_i$ via
  \emph{concatenation} or \emph{cross-modal attention} (\textbf{Q from numeric},
  \textbf{K/V from text}). The LR head predicts Up/Down; evaluation uses Random
  split, Time-based split, and 5-fold TSS.}

    \label{fig:embed_pipeline}
\end{figure}
\vspace{-5pt}

\section{Literature Review}
The integration of ML and NLP for financial analysis has evolved from early approaches combining time-series analysis with news sentiment, which improved prediction accuracy by 12-15\% over purely numerical models \cite{finbert_lstm_2024}. Architectural innovations like DeBERTaV3 further improved performance on benchmarks \cite{he2021debertav3}. Real-time trading demands prompted development of efficient models through techniques like deep self-attention distillation \cite{boosting_transformers_2023}. MiniLM maintained 99\% of base model accuracy with 60\% fewer parameters, while TinyFinBERT leveraged GPT-4 synthetic data to achieve comparable results with 40\% parameter reduction \cite{minilm_distillation_2020}. Recent variants like ModernBERT offer longer context windows and optimized attention mechanisms, though their advantages primarily appear in training efficiency \cite{yarn_context_extension_2023}. Hybrid approaches combining language models with temporal components, such as FinBERT-LSTM architectures, have reduced prediction error by 23\% \cite{finbert_lstm_sp500_2023}. Despite these advances, challenges persist in data quality, interpretability, and adapting to the non-stationary nature of financial markets, \emph{as well as the need for domain-specific evaluation beyond forecasting (e.g., regulatory compliance verification)}~\cite{berger2023towards}. In this paper, we seek to address these challenges by fusing numerical and sentiment embeddings and validating through rolling backtests using various financial metrics.

\section{Methodology} \label{section:methodology}

This section delves into our pipeline for stock movement classification. The pipeline comprises three components: 1) Data Preparation 2) Feature Generation 3) Market Vector Generation. 

\subsection{Data Preparation}

Stock data was collected using \textit{ yfinance} API\footnote{\url{https://ranaroussi.github.io/yfinance}} covering the period from 2007 to 2023, retrieving the \textit{Open}, \textit{High}, \textit{Close}, and \textit{Volume} columns. Daily stock movement was calculated as the difference between the opening price on day \textbf{t} and the closing price on day \textbf{t-1}, as defined by Equation~\eqref{eq:movement}. News articles were preprocessed and subsequently annotated with sentiment scores using the RoBERTa model \cite{liu2019roberta}. Finally, textual and numerical features were merged into a unified dataset. To prevent target leakage in our modeling process, we introduced a one day time lag for all numerical columns except \textit{Open}, ensuring that only historically available information was used for prediction.
\begin{equation}
\mathrm{Movement}_t = P_t^{\mathrm{open}} - P_{t-1}^{\mathrm{close}}
\label{eq:movement}
\end{equation}

\subsection{Feature Generation}
Textual and numerical features are engineered from the merged data set to capture comprehensive feature representations for multimodal classification.

\subsubsection{Textual Embedding Generation}
We used five popular encoder-based models for embedding generation, all sourced from the Hugging Face Model Hub.\footnote{\url{https://huggingface.co/models}}
\begin{itemize}
  \item \textbf{FinBERT}: Financial domain sentiment encoder.
\item \textbf{ModernBERT}: Recent web/news pretraining; longer context.
\item \textbf{Electra}: Generator–discriminator pretraining; efficient.
\item \textbf{DeBERTa}: Disentangled attention; stronger contextualization.
\item \textbf{MiniLM}: Deep self-attention distillation; compact.

\end{itemize}

\subsubsection{Numerical Feature Generation}
Eight numerical features were selected and scaled using the standard scaler to normalize their distributions and ensure consistency.  These features are \textit{Open}, \textit{sentiment volatility}, \textit{aggregate sentiment score}, \textit{Close}, \textit{High}, \textit{Volume}, \textit{Daily Return}, and \textit{Volatility}. 
% They are defined as:
% \[
% \begin{aligned}
% \text{Daily Return}_t &= \frac{\text{Close}_t - \text{Close}_{t-1}}{\text{Close}_{t-1}},\\
% \text{Volatility}_t &= \sqrt{\frac{1}{n-1}\sum_{i=t-n+1}^{t}\bigl(R_i - \overline{R}_t\bigr)^2},\\
% \text{Sentiment Volatility}_t &= \sqrt{\frac{1}{n-1}\sum_{i=t-n+1}^{t}\bigl(s_i - \overline{s}_t\bigr)^2},\\
% \text{Aggregate Sentiment Score}_t &= \frac{1}{m}\sum_{j=1}^{m}s_{t,j}\,.
% \end{aligned}
% \]
% where \(R_i\) are daily returns, \(s_i\) sentiment scores, \(n\) is the rolling-window length, and \(m\) is the number of articles per day.  

% \begin{figure}[ht]
%     \centering
%     \includegraphics[width=5cm]{figures/method_new.png}
%     \caption{ \textbf{Multimodal market prediction pipeline.} Numerical $\left(x_i\right)$ and textual $\left(y_i\right)$ features from each date $\left(d_i\right)$ are transformed into embeddings via feature scaling, $\left(F_s\right)$  and encoder-based models, E, respectively. They are combined into a market vector ( $m_i$ ) using concatenation or cross-modal attention and passed to a logistic regression classifier to predict stock movement (Up/Down).
% }
%     \label{fig:embed_pipeline}
% \end{figure}

\subsection{Market Vector Generation}

Two different methods were used to combine numerical and textual data into unified market vectors.

\subsubsection{\textbf{Concatenation}}
Simple concatenation is used to directly merge scaled numerical features with the previous day's aggregated textual embedding, producing a combined market vector for the day.

\subsubsection{\textbf{Cross-Modal Attention}}
Let $x_i\in\mathbb{R}^{d_n}$ (numeric) and $y_i\in\mathbb{R}^{d_t}$ (text).
Project $X_i=W_x x_i,\; Y_i=W_y y_i$ and compute cross-attention with
queries from $X_i$ and keys/values from $Y_i$:
$Q=X_iW_Q,\; K=Y_iW_K,\; V=Y_iW_V,\;
A_i=\mathrm{softmax}\!\big(QK^\top/\sqrt{d_k}\big)V$ (multi-head, $h$ heads).
The fused market vector $m_i=[X_i;A_i]W_f$ (layer-norm and dropout applied
pre-fusion) feeds a logistic-regression head (Fig.~\ref{fig:embed_pipeline});
MLP and LightGBM baselines overfit and underperformed LR on held-out splits,
consistent with prior reports~\cite{buczynski2023financial,wu2020conditional}.

\section{Datasets} \label{section:datasets}

We used two datasets for this project. The primary FinSen dataset (160k rows; columns: Title, Tag, Content) covers S\&P 500 financial articles from 2007 to 2023\footnote{\url{https://github.com/EagleAdelaide/FinSen_Dataset/tree/main}}. The secondary Financial PhraseBank + FiQA dataset (5,842 rows; columns: Sentence, Sentiment) provides explicit labels for fine-tuning our models\footnote{\url{https://www.kaggle.com/datasets/sbhatti/financial-sentiment-analysis/data?select=data.csv}}.  

\section{Experiments} \label{section:experiments}
\subsection{Dataset Splitting}
To facilitate comparison with related work, we employed three distinct data‐partitioning strategies:
\begin{itemize}
  \item \textbf{Random Split:} The data are randomly shuffled into an 80\% training set and 20\% test set, which breaks the time-series order and introduces information leakage. We retain this split only as a benchmark—its prevalence in the literature facilitates comparison despite its methodological flaws.
  \item \textbf{Time-Based Split:} Observations are ordered chronologically, with the earliest 80\% of dates used for training and the most recent 20\% reserved for testing. This preserves temporal causality, but still evaluates on only a single contiguous test period, which may not capture variability across different market conditions.
  \item \textbf{5-Fold TimeSeriesSplit:} The data is partitioned into five non-overlapping folds along the time axis. In each iteration, fold $k$ (later time segment) serves as the test set, and all preceding folds (earlier segments) form the training set.
\end{itemize}

We use 5-Fold TimeSeriesSplit as our primary validation strategy because random and single time-based splits can leak future information or rely on one test window. Random and single time-based splits—though used in prior work—tend to produce inflated accuracy estimates. This rolling evaluation samples multiple out-of-sample intervals across varying market regimes, yielding more realistic performance metrics and reflects the non-stationary nature of financial data.

% \begin{figure}[ht]
%     \centering
%     \includegraphics[width=7.5cm]{figures/training_vs_testing_years.png}
%     \caption{Training vs Testing Folds.}
%     \label{fig:tss}
% \end{figure}

% \begin{figure}[ht]
%     \centering
%     \includegraphics[width=6.5cm]{figures/Dashboard 1.png}
%     \caption{Fold-Wise Rolling Volatility.}
%     \label{fig:rolling_vol}
% \end{figure}

\subsection{Baseline}
\subsubsection{Logistic Regression}
We used a simple Logistic Regression Classifier on the numerical features without the annotated sentiment and then we added the sentiment score as an additional numerical feature to see how the results changed. 
\subsubsection{Frozen LLM}
To assess the out-of-the-box performance of state-of-the-art language models, we employed QWEN 2.5 7B \cite{yang2024qwen2}, an open-source resource-efficient LLM that excels on financial benchmarks \cite{lin2025open}. We evaluated its capabilities by processing our test data through zero-shot, one-shot, and few-shot prompting paradigms. Throughout all experiments, model parameters remained frozen, allowing us to measure the inherent predictive capabilities without any task-specific fine-tuning. The base prompt template used in our experiments is specified below.

\begin{tcolorbox}[colback=gray!5,colframe=black!50,
                  title=Prompt Template\label{prompt}]
\scriptsize
Based on the numerical data and the news article from yesterday, determine if today’s stock will move 'up' or 'down'. Respond with exactly one word: either 'up' or 'down'. Do not include any additional text or echo any part of this prompt.

\textless Numerical Data\textgreater: {numerical\_data}

\textless News Article\textgreater: {news\_article}

\textless /Answer/\textgreater:
\end{tcolorbox}

\subsection{Evaluation Metrics}
\subsubsection{Classification Metrics}
\begin{itemize}
  \item \textbf{Accuracy:} Fraction of correctly classified days.
  \item \textbf{Precision:} Proportion of predicted up‐days that were actually up.
  \item \textbf{Recall:} Proportion of actual up‐days correctly identified.
  \item \textbf{F‐score:} Harmonic mean of precision and recall.

\end{itemize}

\subsubsection{Financial Metrics}
For our long-only trading strategy (positions \(p_t\in\{0,1\}\)), we evaluate the following performance metrics:

\begin{itemize}
  \item \textbf{MCC:} Matthews correlation coefficient for trading signals, quantifying alignment between predicted positions and actual profits under imbalanced outcomes.  
  \item \textbf{Directional Win Rate (DWR):} Proportion of long positions yielding a positive return.  
    \[
      \mathrm{DWR} = \frac{\sum_{t: p_t\,r_t > 0} 1}{\sum_{t: p_t = 1} 1}
    \]
    where \(r_t\) are the daily returns and \(p_t\) the long‐only positions.
  \item \textbf{Profit Factor (PF):} Ratio of gross profits to gross losses.  
    \[
      \mathrm{PF} = \frac{\sum_{t: p_t\,r_t > 0} p_t\,r_t}{-\sum_{t: p_t\,r_t < 0} p_t\,r_t}
    \]
  \item \textbf{Sharpe Ratio (SR):} Risk‐adjusted return, computed as the mean strategy return divided by its standard deviation (annualized factor omitted for simplicity).  
    \[
      \mathrm{SR} = \frac{\sum_{t} p_t\,r_t \;/\;\sum_{t}1}{\sqrt{\sum_{t}(p_t\,r_t - \overline{p_t\,r_t})^2 \;/\;\sum_{t}1}}
    \]
\end{itemize}

\section{Results}

\subsubsection{Baseline Results}

The results in Tables \ref{tab:1} and \ref{tab:2} demonstrate that incorporating sentiment scores alongside numerical features—and applying SMOTE balancing—consistently boosts performance across both classification and financial metrics under 5-fold TimeSeriesSplit. Focusing on our primary validation (Tables \ref{tab:4} and \ref{tab:5}), \emph{Concatenation} fusion with \textbf{MiniLM} delivers a compelling combination of predictive accuracy (0.65) and F1 (0.72), as well as the highest profitability (Profit Factor 2.03, Sharpe 3.15). Meanwhile, \emph{Cross-Modal Attention} fusion with \textbf{DeBERTa} achieves the top classification results (accuracy 0.68, F1 0.73) while maintaining robust financial returns (Profit Factor 1.75, Sharpe 2.24).  

These observations motivate our fine-tuning choices: we refine MiniLM in the concatenation pipeline to further amplify its high-risk-adjusted returns, and we fine-tune DeBERTa in the attention pipeline to capitalize on its superior contextual representation and classification strength.

% \usepackage{graphicx}
\begin{table}[htbp]
\caption{Logistic Regression (LR) Baseline \\ \textit{Table presents the \textbf{classification} metrics using LR. Best results are highlighted in bold.}}
\label{tab:1}
\resizebox{\columnwidth}{!}{%
\begin{tabular}{llcccc}
\hline
Features                                                                & Configuration & Accuracy & Precision & Recall & F1 score \\ \hline
Numerical                                                               &               &          &           &        &          \\
                                                                        & LR            & 0.55     & 0.45      & 0.59   & 0.45     \\
                                                                        & LR + SMOTE    & \textbf{0.56}     & \textbf{0.46}      & \textbf{0.67}   & \textbf{0.53}     \\ \hline
\begin{tabular}[c]{@{}l@{}}Sentiment score \\ \& Numerical\end{tabular} &               &          &           &        &          \\
                                                                        & LR            & 0.62     & 0.66      & \textbf{0.86}   & 0.70     \\
                                                                        & LR + SMOTE    & \textbf{0.65}     & \textbf{0.68}      & 0.81   & \textbf{0.71}     \\ \hline
\end{tabular}%
}
\end{table}

% \usepackage{graphicx}
\begin{table}[htbp]
\caption{Logistic Regression (LR) Baseline \\ \textit{Table presents the \textbf{financial} metrics using LR. Best results are highlighted in bold.}}

\label{tab:2}
\resizebox{\columnwidth}{!}{%
\begin{tabular}{llllll}
\hline
Features                                                                & Configuration & MCC  & \begin{tabular}[c]{@{}l@{}}Directional \\ Win Rate\end{tabular} & Profit Factor & Sharpe Ratio \\ \hline
Numerical                                                               &               &      &                                                                 &               &              \\
                                                                        & LR            & \textbf{0.06} & 0.31                                                            & \textbf{1.22}          & 0.79         \\
                                                                        & LR + SMOTE    & \textbf{0.06} & \textbf{0.35}                                                            & \textbf{1.22}          & \textbf{0.84}         \\ \hline
\begin{tabular}[c]{@{}l@{}}Sentiment Score \\ \& Numerical\end{tabular} &               &      &                                                                 &               &              \\
                                                                        & LR            & 0.27 & \textbf{0.45}                                                            & 1.57          & 2.14         \\
                                                                        & LR + SMOTE    & \textbf{0.35} & 0.42                                                            & \textbf{1.91}          & \textbf{2.86}         \\ \hline
\end{tabular}%
}
\end{table}
% \usepackage{graphicx}
% \usepackage{graphicx}
% \usepackage{graphicx}
\begin{table}[]
\caption{LLM Inference \\ \textit{Table presents the \textbf{classification} \& \textbf{financial} metrics for Frozen LLM. Best results are highlighted in bold.}}
\label{tab:3}
\resizebox{\columnwidth}{!}{%
\begin{tabular}{lllllllll}
\hline
Prompting Paradigm & Accuracy      & Precision     & Recall        & F1            & MCC           & DWR           & PF            & SR            \\ \hline
Zero-Shot          & 0.51          & \textbf{0.60} & 0.42          & 0.50          & \textbf{0.05} & 0.22          & \textbf{1.16}          & 0.47          \\
1-Shot             & \textbf{0.56} & 0.57          & 0.90          & \textbf{0.70} & 0.04          & 0.49          & \textbf{1.16} & \textbf{0.69} \\
Few-Shot           & 0.55          & 0.56          & \textbf{0.95} & \textbf{0.70}          & -0.04         & \textbf{0.51} & 1.14          & 0.63          \\ \hline
\end{tabular}%
}
\end{table}

\subsubsection{FineTuning Results}  
After fine‐tuning on the FiQA+Financial PhraseBank corpus, both fusion pipelines yield measurable improvements in downstream stock‐movement classification:

\begin{itemize}
  \item \textbf{MiniLM (Concatenation Fusion):} Acc $0.65\!\to\!0.67$ and F1 $0.72\!\to\!0.75$ (Table~\ref{tab:4}); MCC $0.27\!\to\!0.31$ (Table~\ref{tab:5}), while PF declines $2.03\!\to\!1.72$ due to a tighter post–fine-tuning decision boundary that increases false-positive longs; the attention-based DeBERTa variant avoids this trade-off (SR $2.55$).

  \item \textbf{DeBERTa (Cross-Modal Attention):} F1 $0.73\!\to\!0.75$ (Table~\ref{tab:4}), MCC $0.32\!\to\!0.33$, and SR $2.24\!\to\!2.55$ (Table~\ref{tab:5}); fine-tuning sharpens attention-head sensitivity to domain-specific linguistic cues, yielding concurrent gains in classification and risk-adjusted return.
\end{itemize}

These gains arise from fine-tuned domain signals, multi-head feature discrimination, and finer-grained pattern capture for robust predictions.

\subsubsection{LLM Inference}

Among the three prompting paradigms, one-shot prompting appears to yield the most balanced classification and financial metrics (Table \ref{tab:3}), but this likely reflects memorization rather than true forecasting. As Lopez-Lira et al. \cite{lopez-lira2025memorization} show, LLMs can recall pretraining headlines when given minimal cues. Providing a single example thus unlocks memorized patterns—artificially boosting recall—without ensuring out-of-sample generalization. Consequently, one-shot recall gains should be viewed cautiously, as they may stem from data leakage rather than genuine inference.

% \usepackage{graphicx}
% \usepackage{graphicx}
\begin{table}[]
\caption{ Encoder Based Models Evaluation \\ \textit{Table presents the \textbf{classification} metrics. Best results are highlighted in bold.}}
\label{tab:4}
\resizebox{\columnwidth}{!}{%
\begin{tabular}{llllll}
\hline
\begin{tabular}[c]{@{}l@{}}Market Vector \\ Generation\end{tabular} & Model             & Accuracy      & Precision     & Recall        & F1 score      \\ \hline
Concatenation                                                       &                   &               &               &               &               \\
                                                                    & FinBERT           & 0.56          & 0.60          & 0.68          & 0.63          \\
                                                                    & ModernBERT        & 0.54          & 0.63          & 0.46          & 0.51          \\
                                                                    & DeBERTa          & 0.58          & 0.63          & 0.61          & 0.60          \\
                                                                    & Electra           & 0.55          & 0.60          & 0.59          & 0.59          \\
                                                                    & MiniLM            & 0.65          & \textbf{0.65}          & 0.81          & 0.72          \\
                                                                    & MiniLM- FineTuned & \textbf{0.67} & \textbf{0.65} & \textbf{0.88} & \textbf{0.75} \\ \hline
Attention                                                           &                   &               &               &               &               \\
                                                                    & FinBERT           & 0.59          & 0.65          & 0.59          & 0.61          \\
                                                                    & ModernBERT        & 0.63          & 0.67          & 0.66          & 0.65          \\
                                                                    & DeBERTa          & \textbf{0.68} & 0.70          & \textbf{0.77} & \textbf{0.73} \\
                                                                    & Electra           & 0.62          & 0.67          & 0.62          & 0.62          \\
                                                                    & MiniLM            & 0.65          & 0.67          & 0.73          & 0.69          \\
                                                                    & DeBERTa-FineTuned & 0.67          & \textbf{0.71} & 0.71          & 0.70          \\ \hline
\end{tabular}%
}
\end{table}

% \usepackage{graphicx}
% \usepackage{graphicx}
% \usepackage{graphicx}
\begin{table}[]
\caption{Encoder Based Models Evaluation \\ \textit{Table presents the \textbf{financial} metrics. Best results are highlighted in bold.}}
\label{tab:5}
\resizebox{\columnwidth}{!}{%
\begin{tabular}{llllll}
\hline
\begin{tabular}[c]{@{}l@{}}Market Vector \\ Generation\end{tabular} & Model             & MCC           & \begin{tabular}[c]{@{}l@{}}Directional \\ Win Rate\end{tabular} & Profit Factor & Sharpe Ratio  \\ \hline
Concatenation                                                       &                   &               &                                                                 &               &               \\
                                                                    & FinBERT           & 0.12          & 0.35                                                            & 1.42          & 1.52          \\
                                                                    & ModernBERT        & 0.11          & 0.23                                                            & 1.47          & 1.20          \\
                                                                    & DeBERTa          & 0.16          & 0.30                                                            & 1.54          & 1.64          \\
                                                                    & Electra           & 0.09          & 0.30                                                            & 1.35          & 1.17          \\
                                                                    & MiniLM            & 0.27          & 0.42                                                            & \textbf{2.03} & \textbf{3.15} \\
                                                                    & MiniLM- FineTuned & \textbf{0.31} & \textbf{0.45}                                                   & 1.72          & 2.57          \\ \hline
Attention                                                           &                   &               &                                                                 &               &               \\
                                                                    & FinBERT           & 0.20          & 0.29                                                            & 1.45          & 1.39          \\
                                                                    & ModernBERT        & 0.23          & 0.32                                                            & 1.60          & 1.77          \\
                                                                    & DeBERTa          & 0.32          & \textbf{0.38}                                                   & 1.75          & 2.24          \\
                                                                    & Electra           & 0.23          & 0.30                                                            & 1.60          & 1.78          \\
                                                                    & MiniLM            & 0.27          & 0.36                                                            & 1.59          & 1.89          \\
                                                                    & DeBERTa-FineTuned & \textbf{0.33} & 0.34                                                            & \textbf{1.88} & \textbf{2.55} \\ \hline
\end{tabular}%
}
\end{table}

\section{Conclusion and Future Work}
In this study, we presented STONK, a multimodal framework integrating numerical market indicators with sentiment-enriched news embeddings via feature concatenation and cross-modal attention. Backtesting results demonstrate significant gains in predictive accuracy and risk-adjusted returns over numeric-only baselines, confirming the value of multimodal integration. Future work will address live trading deployment, multi-asset portfolio extension, additional data modalities, and advanced fusion and continual learning techniques to enhance adaptability in dynamic markets.\label{section:conclusion}

\section*{Acknowledgment}
Portions of the text across all sections were refined with assistance from GPT-4o \cite{openai2024gpt4ocard}.

% \section{Acknowledgment}

% This research has been funded by the Federal Ministry of Education and Research of Germany and the state of North-Rhine Westphalia as part of the Lamarr-Institute for Machine Learning and Artificial Intelligence, LAMARR22B.

% \renewcommand*{\bibfont}{\footnotesize}
% \printbibliography

\bibliographystyle{IEEEtran}
\bibliography{bibliography}

\end{document}